\documentclass[sigconf]{acmart}
\usepackage{booktabs, multirow}
\usepackage{amsmath}
\usepackage{xcolor}
\usepackage{xspace}
\usepackage{tcolorbox}
\usepackage[table]{xcolor}
\usepackage{subfigure}
\usepackage{fix-cm}
\usepackage{enumitem}
\usepackage{setspace}
\usepackage{tikz}
\usetikzlibrary{tikzmark}
\usepackage{float}
\usepackage{dsfont}
\usepackage{pifont}
\usepackage{bbding}
\usepackage{graphicx}
\usepackage{subfigure}
\usepackage{subcaption}

\usepackage{color}
\usepackage{tikz}
\usepackage[edges]{forest}
\usepackage{tcolorbox}
\usepackage[table]{xcolor}

\definecolor{hidden-draw}{RGB}{0,0,0}
\definecolor{hidden-pink}{rgb}{0.98, 0.94, 0.75}
\definecolor{level0}{rgb}{0.67, 0.88, 0.69}
\definecolor{level1}{rgb}{0.98, 0.92, 0.84}
\definecolor{level2}{rgb}{0.8, 0.8, 1.0}
\definecolor{level3}{rgb}{1.0, 0.71, 0.76}

\definecolor{yellow}{rgb}{0.74, 0.56, 0}
\definecolor{purple}{rgb}{0.32, 0.09, 0.98} 
\definecolor{red}{rgb}{0.81, 0.09, 0.13}
\definecolor{mycolor}{rgb}{0.18,0.29,0.79}

\definecolor{blue}{cmyk}{0.95,0.0,0.2,0.2}
\definecolor{lightyellow}{cmyk}{0.01,0.0,0.2,0.01}
\definecolor{lightblue}{cmyk}{0.1,0.0,0.02,0.02}
\definecolor{viridis3}{RGB}{33, 145, 140} 
\definecolor{mypurple}{RGB}{236, 203, 250}

\newcommand{\cubeicon}{%
\raisebox{-0.4ex}{%
\begin{tikzpicture}[scale=0.2]
  \filldraw[fill=gray!20] (0,0) rectangle (1,1);
  \filldraw[fill=gray!30] (1,0) -- (1.3,0.3) -- (1.3,1.3) -- (1,1) -- cycle;
  \filldraw[fill=gray!15] (0,1) -- (0.3,1.3) -- (1.3,1.3) -- (1,1) -- cycle;
\end{tikzpicture}}}

\usepackage[normalem]{ulem}
\newcommand{\lb}{$\langle$}
\newcommand{\rb}{$\rangle$}

\tikzset{mycircled/.style={circle,draw=black,inner sep=0.1em,line width=0.1em, scale=0.8}} 

\newcommand{\mcrag}{\texttt{MultiCube-RAG}\xspace}

\newcommand{\ul}{\underline}

\newcommand{\nop}[1]{}

\AtBeginDocument{%
  }

\setcopyright{acmlicensed}
\copyrightyear{2025}
\acmYear{2025}
\acmDOI{XXXXXXX.XXXXXXX}
\acmConference[Conference acronym 'XX]{Make sure to enter the correct
  conference title from your rights confirmation email}{June 03--05,
  2018}{Woodstock, NY}

\begin{document}

\title{MultiCube-RAG for Multi-hop Question Answering}


\author{
Jimeng Shi$^{1}$, Wei Hu$^1$, Runchu Tian$^1$, Bowen Jin$^1$, Wonbin Kweon$^1$, SeongKu Kang$^2$, \\ Yunfan Kang$^1$, Dingqi Ye$^1$, Sizhe Zhou$^1$, Shaowen Wang$^1$, Jiawei Han$^1$
}
\affiliation{%
\institution{$^1$University of Illinois Urbana-Champaign,  $^2$Korea University}
  \country{}
}
\email{
{jimeng8, shaowen, hanj}@illinois.edu
}

\renewcommand{\shortauthors}{Jimeng Shi et al.}

\begin{abstract}
Multi-hop question answering (QA) necessitates multi-step reasoning and retrieval across interconnected subjects, attributes, and relations.
Existing retrieval-augmented generation (RAG) methods struggle to 
capture these structural semantics accurately, resulting in suboptimal performance.
Graph-based RAGs structure such information in graphs, but the resulting graphs are often noisy and computationally expensive.
Moreover, most methods rely on single-step retrieval, neglecting the need for multi-hop reasoning processes.
Recent training-based approaches attempt to incentivize the large language models (LLMs) for iterative reasoning and retrieval, but their training processes are prone to unstable convergence and high computational overhead.
To address these limitations, we devise an \textit{ontology-based cube structure} with multiple and orthogonal dimensions to model structural subjects, attributes, and relations.
Built on the cube structure, we propose \textbf{MultiCube-RAG}, a \textit{training-free method consisting of multiple cubes for multi-step reasoning and retrieval}. 
Each cube specializes in modeling a class of subjects, so that \mcrag flexibly selects the most suitable cubes to acquire the relevant knowledge precisely.
To enhance the query-based reasoning and retrieval, our method decomposes a complex multi-hop query into a set of simple subqueries along cube dimensions and conquers each of them sequentially.
Experiments on four multi-hop QA datasets show that \mcrag improves response accuracy by 8.9\% over the average performance of various baselines. 
Notably, we also demonstrate that our method performs with greater efficiency and inherent explainability. 
The code is available at \textcolor{mycolor}{\emph{\url{https://anonymous.4open.science/r/CubeRAG/}}}. 
\end{abstract}

\maketitle

\begin{figure}[t]
\centering
  \includegraphics[width=\columnwidth]{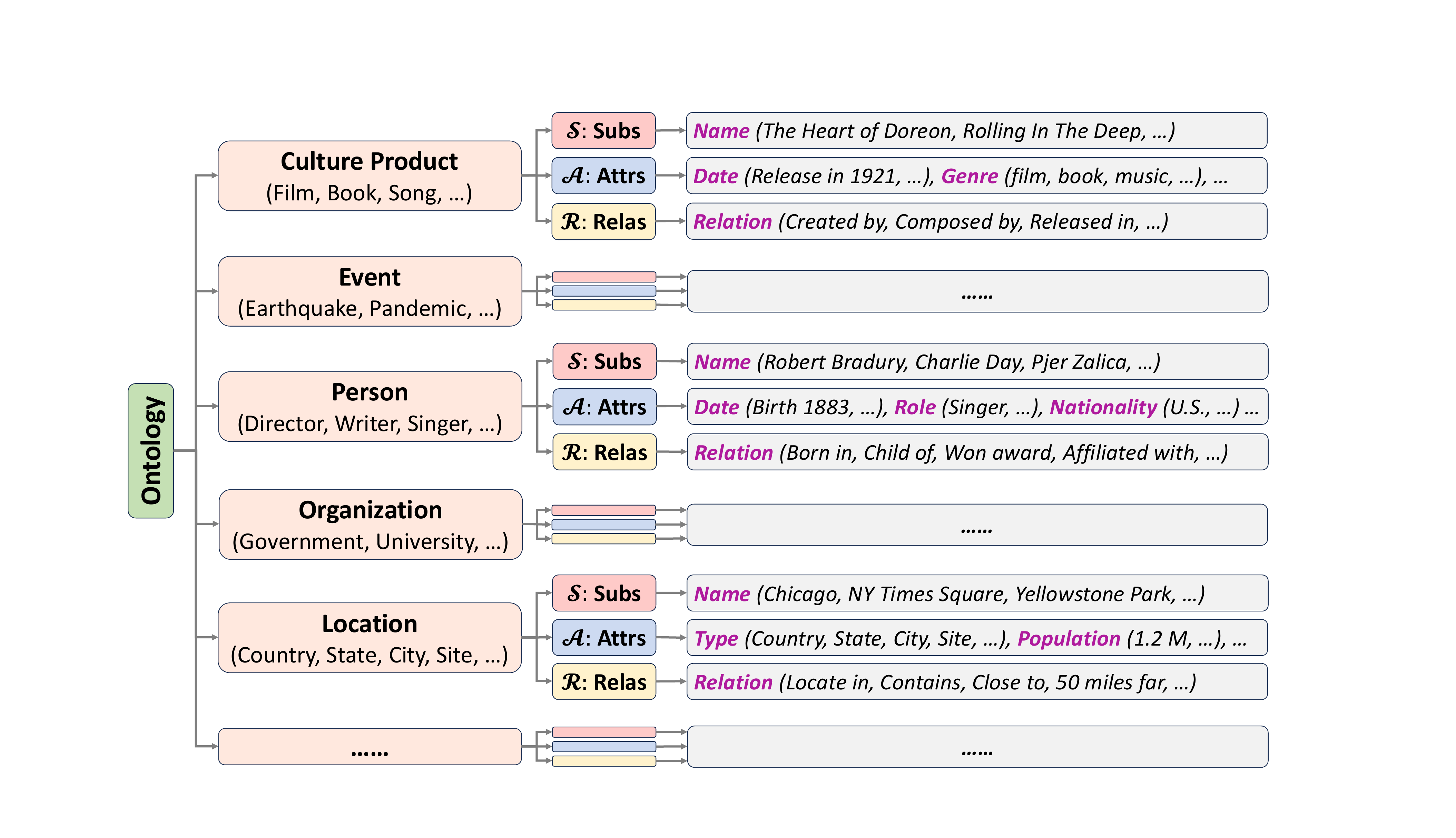}
  \caption{An example of an ontology structure extracted by the LLM.  
  $\mathcal{S}, \mathcal{A}, \mathcal{R}$ denote a set of subjects, attributes, and relations, respectively.
  }
  \vspace{-3mm}
  \label{fig:ontology_cube}
\end{figure}

\section{Introduction}
\label{sec:intro}
While large language models (LLMs) have achieved notable success in natural language understanding and a wide range of question-answering tasks \cite{clark2018think, karanikolas2023large}, they often struggle with complex \textit{multi-hop} queries that require multi-step reasoning and knowledge aggregation \cite{min2019multi, yang2018hotpotqa}. 
Chain-of-thought (CoT) has been developed to mitigate this challenge by prompting LLMs to generate a reasoning trajectory \cite{wei2022chain, yu2021multi}. 
Despite the advancements, the fundamental problem of hallucination persists due to the lack of up-to-date knowledge during pre-training \cite{huang2024coq, augenstein2024factuality}, necessitating the
external acquisition.
Therefore, integrating \textit{multi-step reasoning and retrieval} is essential for effectively solving multi-hop queries.

Existing iterative reasoning and retrieval methods are typically implemented through prompting, supervised training, or reinforcement learning (RL)-based training.
Prompting-based methods~\cite{trivedi2022interleaving, shao2023enhancing} elicit step-by-step reasoning chains by prompts, which are poorly adaptable.
Training-based methods \cite{schick2023toolformer, li2025beyond, asai2024self} offer greater adaptability but are constrained by the need for large-scale, high-quality annotated trajectories, exhibiting less effectiveness for scalability.
Recently, reinforcement learning (RL) (a.k.a. R1-based) methods \cite{jin2025search, song2025r1, chen2025learning} allow LLMs to learn multi-turn behaviors via outcome rewards. 
However, these RL approaches face significant hurdles, such as unstable training dynamics \cite{xue2025simpletir, wang2025ragen, baronio2025kevin}, complex reward engineering \cite{kwon2023reward, gao2024designing}, and prohibitive computational overhead \cite{jin2025search, song2025r1}.
Moreover, retrievers themselves (e.g., sparse \cite{kadhim2019term}, dense \cite{karpukhin2020dense, kang2024improving}) struggle to leverage structural subjects, attributes, and relations in the text data, which may lead to suboptimal retrieval, which subsequently degrades the reasoning process. 
Graph-based retrievers \cite{edge2024local, guo2024lightrag, gutierrez2024hipporag} preserve structural information, but they often suffer from semantic noise by modeling exhaustive details. Many attributes (e.g., a location’s time zone or capital) are redundant, as they represent general knowledge already internalized within an LLM's prior knowledge.
Moreover, the resulting graph density leads to computational bottlenecks in both construction and multi-hop reasoning, particularly for large corpora~\cite{huang2025survey}.

In this study, we address these limitations through two aspects: \textbf{a succinct cube structure} and \textbf{a cube-based RAG framework}.
First, we design an ontology-guided structure with multiple orthogonal dimensions (i.e., cube) to comprehensively and concisely model the structural semantics of subjects ($\mathcal{S}$), attributes ($\mathcal{A}$), and relations ($\mathcal{R}$) in a corpus, as shown in Figure \ref{fig:ontology_cube}. 
Rather than treating documents as flat text or dense graphs, we map subjects, attributes, and relations onto distinct dimensions of a single or multiple cubes.
Such orthogonality across multiple dimensions facilitates a comprehensive and concise structural representation of documents.
A document (or a passage) typically describes one or a small number of entities/events (e.g., \textit{hurricane}), their attributes (e.g., \textit{time, location, intensity}), and relations with other entities/events (e.g., \textit{landed at, caused loss, reported by}). 
Multiple documents representing the same or similar kind of entities/events (e.g., \textit{hurricanes, storms, tornadoes}) can be grouped together as ``weather events'' to stay in the same cube, with the identity of events/entities (e.g., ``Event name'', ``Event type'') as the main cube dimension, and some of their associated properties/relations (e.g., ``Date'', ``Location'', ``Person (Reporter)'', \ldots) can be linked to other dimensions. 
Given an ontology structure \cite{zulkipli2022systematic}, a multi-cube model can be used to structure any kind of document/subject in a corpus.

\begin{figure}[ht!]
\centering
  \includegraphics[width=0.99\columnwidth]{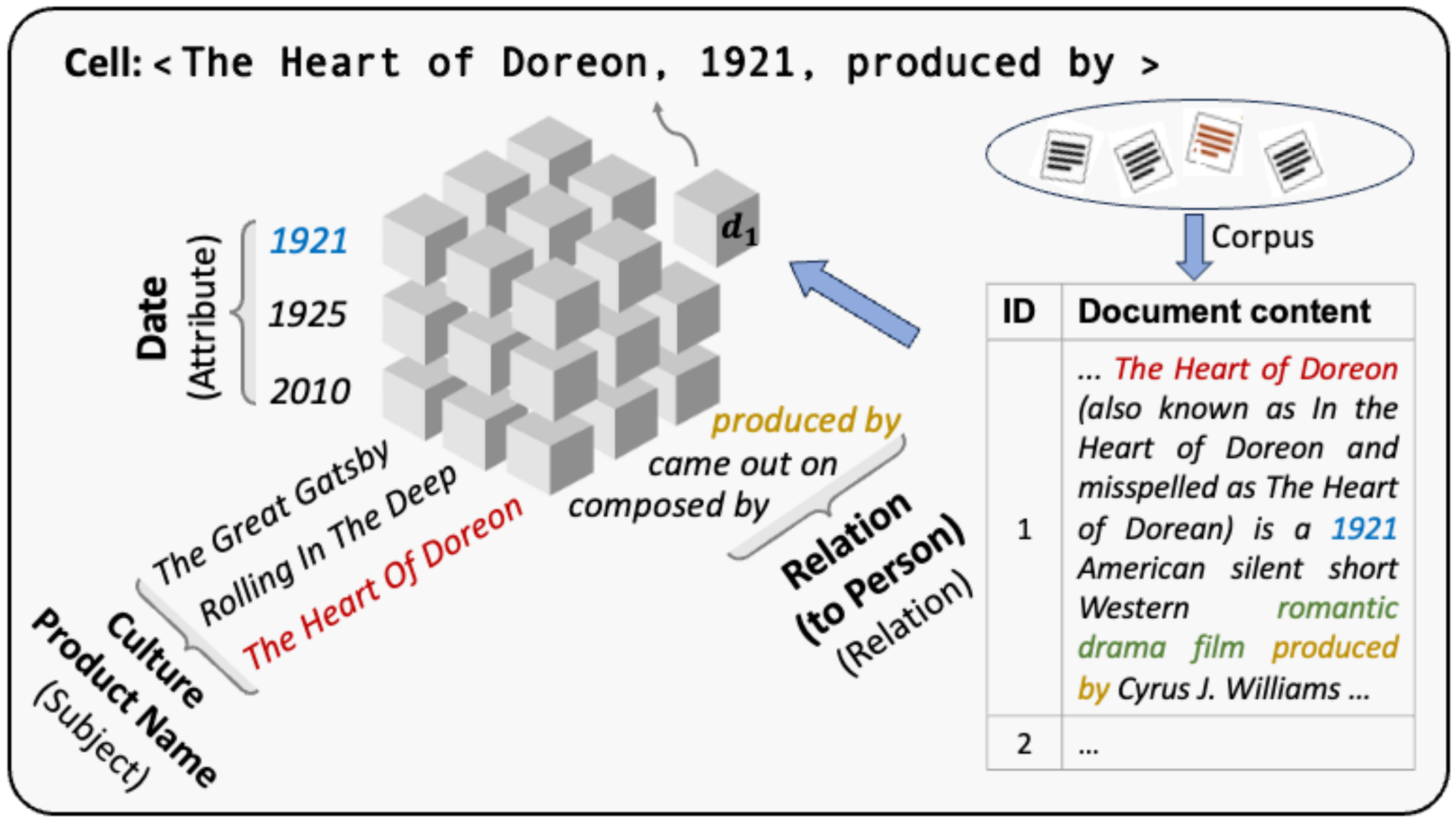}
  \caption{An ontology-guided cube (``Culture Product'') and document allocation into cube cell. We show only one of the attributes (i.e., release date) here; more attributes (e.g., genre, rating), and relations to locations (e.g., filmed at, release at) are included as other cube dimensions but are not shown up.
  Different dimensions are highlighted with distinct colors.
  } 
  \label{fig:product_cube}
\end{figure}

\noindent
\textbf{Example 1.}
In Fig. \ref{fig:product_cube}, a document describing a cultural product naturally organizes around subjects (e.g., film), their attributes (e.g., release date, genre), and relational links (e.g., produced by, written by). 
Specifically, document $d_1$ is represented with a cube\footnote{Unless otherwise specified, the term ``cube'' refers to a large ``hypercube'' with multiple dimensions.}
across multiple dimensions $d_1 \rightarrow$ \cubeicon\ [``Culture Product Name'': \textit{``The Heart Of Doreon''}, ``Date'': \textit{``1921''}, ``Genre'': \textit{``romantic drama film''}, ``Relation (to Person)'': \textit{``produced by''},  \ldots].
Noted that each component of $\{\mathcal{S}, \mathcal{A}, \mathcal{R}$\} is not restricted to a single axis; for instance, the attribute set for a film (e.g., release date, genre, rating) can correspond to multiple independent dimensions, although only one of the attributes (release date) is shown as a single axis for the sake of 3D visualization.


With the cube structure established, a cube can be constructed by projecting documents into their corresponding cube cells. 
Given the designed structure (cube dimensions), an LLM can be used to go over all the documents in a corpus to
extract the instance values of the corresponding dimensions, and then link the document IDs in the corresponding cube cells whose dimensional values exist (as shown in Example 1).
The resulting cube cells act as concise ``descriptors'' of the underlying document content, enabling 
efficient search and reasoning for the subsequent retrieval tasks.

Built on the constructed multiple cubes, we further propose \mcrag, a retrieval and reasoning framework, tailored to process multi-hop queries.
Given the inherent complexity of multi-hop questions, resolving such queries typically requires an iterative process of reasoning and retrieval to infer intermediate entities before identifying the final answer \cite{trivedi2022interleaving}. 
To this end, \mcrag first decomposes the multi-hop query into a single-hop sub-query as the starting point. It then flexibly selects the most suitable cube to retrieve relevant knowledge for this sub-query to generate an intermediate answer. 
By grounding the subsequent reasoning steps in intermediate results, the model generates the next sub-query, thereby maintaining an iterative retrieval and reasoning loop 
until the final solution is reached.
Fig. \ref{fig:framework} walks through the pipeline with a concrete example.
Sec. \ref{sec:method} details the design of cubes, their associated dimensions, and the \mcrag framework.


We observe three benefits from the above design: 
(1) Ontology-guided cubes leverage the structural semantics across multiple orthogonal dimensions, making the retrieval process effective, efficient, and inherently explainable;
(2) Subject-specific cubes facilitate flexible reasoning and query-dependent retriever selection; and
(3) \mcrag is an iterative framework, significantly enhancing both reasoning depth and retrieval accuracy.
Overall, our contributions are summarized as follows:
\begin{itemize}[leftmargin=0.6cm,nosep]
    \item We hypothesize and demonstrate that a structured and well-planned retrieval system can address the complex multi-hop QA challenge, even without training efforts.
    \item We introduce an ontology-guided cube structure, and propose \mcrag, a framework characterized by iterative reasoning and retrieval, and flexibly selecting the suitable retriever at each iteration of retrieval.    
    \item Experiments across four multi-hop QA datasets prove that our \mcrag exhibits remarkable performance against different baseline approaches.
\end{itemize}

\begin{figure*}[t]
\centering
  \includegraphics[width=0.95\textwidth]{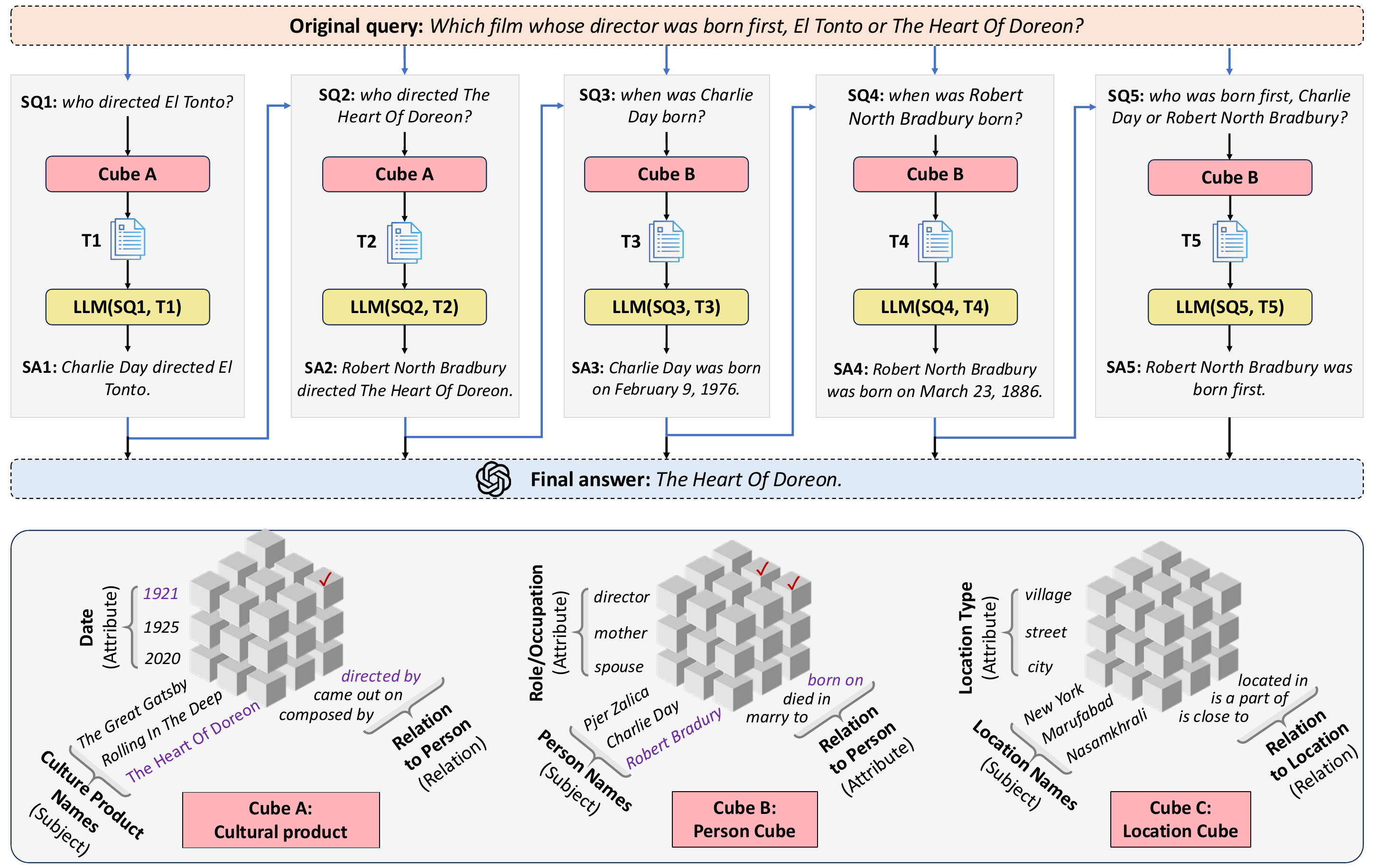}
  \vspace{-2mm}
  \caption{\mcrag for multi-step reasoning and retrieval. It includes multiple ontology-guided cubes (A, B, C) and supports selecting the most suitable one for retrieval at each iteration.
  The dimensional values serve as search keys; the cube cells with red checkmarks are activated using the search keys in purple.
   While the ``Location Cube'' is not used in this example, it is useful for other queries. 
  For abbreviation, SQ: subquery, SA: subquery answer, \#: iteration number, and T: retrieved documents.}
  \label{fig:multicube}
\end{figure*}

\section{Related Work}
\label{sec:related}

\subsection{Retrieval-Augmented Generation}
Retrieval-augmented generation (RAG) can improve LLM accuracy by fetching external context \cite{lewis2020retrieval}. While sparse methods (e.g., BM25 \cite{robertson1994some, trotman2014improvements}) are efficient and interpretable, they lack semantic understanding and perform poorly for paraphrasing. 
Conversely, dense retrievers (e.g., DPR \cite{karpukhin2020dense}, Contriever \cite{izacard2022contriever}) capture nuanced semantics but often sacrifice precision and interpretability \cite{ji2019visual}. Moreover, such methods often neglect the structural information, which is crucial for accurate retrieval, resulting in suboptimal performance.
Graph-based RAG (e.g., GraphRAG \cite{edge2024local}, LightRAG \cite{guo2024lightrag}, HippoRAG \cite{gutierrez2024hipporag, gutierrez2025rag}) provides structural awareness through entity/document graphs. However, such graph-based RAG systems often rely on LLM-generated graphs, which can be noisy, hallucinated, and expensive to construct \cite{huang2025survey}.
Furthermore, most existing approaches employ a single-turn paradigm, making it challenging for multi-hop queries that require iterative reasoning and retrieval aggregation.

\subsection{Multi-step Reasoning and Retrieval}
Iterative reasoning and retrieval typically fall into three paradigms: prompting, supervised fine-tuning (SFT), or reinforcement learning (RL). 
Prompting-based methods (e.g., IRCoT \cite{trivedi2022interleaving}, Iter-RetGen~\cite{shao2023enhancing}) synergize retrieval and generation through Chain-of-Thought (CoT) steps. Similarly, Search-o1 \cite{li2025search} employs an agentic ``Reason-in-Docs'' module to integrate retrieved knowledge into the reasoning process. However, the prompts used are usually poorly generalized and adaptable.
Supervised approaches \cite{schick2023toolformer, li2025beyond, asai2024self} offer adaptability but are bottlenecked by the requirement for high-quality, large-scale annotated trajectories. 
To mitigate this, RL-based frameworks, such as Search-R1 \cite{jin2025search}, R-Searcher \cite{song2025r1}, and R-Search \cite{zhao2025r}, train a single LLM to optimize reasoning-search trajectories via specialized reward functions. 
Notably, s3 \cite{jiang2025s3} decouples these components, training only the searcher using a ``Gain Beyond RAG'' reward.
Despite their success, RL-based models often suffer from erratic training dynamics \cite{xue2025simpletir, baronio2025kevin} and prohibitive computational costs, requiring hundreds of GPU hours even for smaller models Qwen2.5-3B or Qwen2.5-7B \cite{jin2025search}. In this work, we aim to perform multi-step reasoning and retrieval without training.

\subsection{Data Cube and Text Cube}
Data warehouses store multidimensional data organized around specific subjects, commonly referred to as ``data cubes'' \cite{han2022data, harinarayan1996implementing, dyreson1996information}.
Extending this paradigm to unstructured data, text cubes were developed to manage and process textual corpora \cite{lin2008text}.
Considering user reviews of products, they may be evaluated on four dimensions: product model, price, time, and score. Each row stores a user review (i.e., a document or a passage), with values specified for these four dimensions.
Subsequent research has expanded the utility of text cubes; STREAMCUBE \cite{feng2015streamcube} and the Semantics-Space-Time Cube \cite{li2018semantics} incorporate spatial and temporal hierarchies for event clustering, while Doc2Cube \cite{tao2018doc2cube} automates the allocation of documents into cube structures for multidimensional analytics.
Beyond general text mining, cube representations have also been studied in geospatial analysis \cite{wang2023geospatial}, text classification \cite{zhang2023effective}, and topic discovery \cite{zhang2025teleclass}.
More recently, Hypercube-RAG~\cite{shi2025hypercube} applied the cube-based RAG for single-hop queries; it faces significant limitations in the context of solving multi-hop queries, which demands iterative, multi-step reasoning and retrieval. That is the study score in this work.

\section{Methodology}
\label{sec:method}
This section describes the ontology-guided cube design (Sec. \ref{sec:cube_design}) and cube construction (Sec. \ref{sec:cube_construct}).
Leveraging this structure,
we introduce the cube-based retrieval mechanism (Sec. \ref{sec:cube_rag}) and
present \mcrag, a multi-cube-based RAG framework (Sec. \ref{sec:mcrag}). The core modules are illustrated, spanning query decomposition for subquery reasoning, flexible cube selection for retrieval, knowledge aggregation, and iterative loop execution that culminates in the final answer generation.

\subsection{Ontology-Guided Cube Design}
\label{sec:cube_design}
\textbf{Ontology extraction.} We prompt an LLM to extract information from documents for an 
ontology structure as follows: 
``Ontology $\rightarrow$ Subject Classes $\rightarrow$ Attribute/Relation Types $\rightarrow$ Instance Values".
Let the ontology $\mathcal{O}$ be defined as a set of semantic classes $\mathcal{C}$:
\begin{equation}
    \mathcal{O} = \{C_1, C_2, C_3, \dots\},
\end{equation}
For each semantic class $C \in \mathcal{O}$, we define a structural primitive schema $\Phi_C$ as a 3-tuple of sets:
\begin{equation}
    \Phi_C = (\mathcal{S}_C, \mathcal{A}_C, \mathcal{R}_C),
\end{equation}
where:
\begin{equation}
\begin{aligned}
    \mathcal{S}_C &= \{s_{C,1}, s_{C,2}, \dots, s_{C,J}\}, \\
    \mathcal{A}_C &= \{A_C^{(1)}, A_C^{(2)}, \dots, A_C^{(M)}\}, \\
    \mathcal{R}_C &= \{R_C^{(1)}, R_C^{(2)}, \dots, R_C^{(N)}\},
\end{aligned}
\end{equation}
where 
Subjects $\mathcal{S}_C$: a set of entity/event instances belonging to that class (e.g., \textit{Culture Product, Person, Location, \ldots});
Attributes $\mathcal{A}_C$: a set of attribute types (e.g., \textit{Genre, Date, Rating, \ldots}), and each type $A_C^{(m)}=\{a_C^{(1)}, a_C^{(2)}, \dots, a_C^{(p)}\}$ represent a set of attribute instances;
and Relations $\mathcal{R}_C$: a set of relation types (e.g., \textit{links to Person, Location, Organization, \ldots}), and each type $R_C^{(n)}=\{r_C^{(1)}, r_C^{(2)}, \dots, r_C^{(q)}\}$ represent a set of relations instances. 

\noindent
\textbf{Ontology-guided Cube Structure Design.} 
Given an ontology structure, one can generate a set of cube structures based on available data and applications.  
In general, 
the number of cubes and their corresponding types can be determined by the semantic classes $C$ defined in the top levels of the ontology.
Each semantic class $C \in \mathcal{O}$ results in the instantiation of a unique cube $\mathbb{K}_C$.
In many cases, classes sharing highly similar structural schemas can be defined as a single cube to reduce sparsity and improve scalability (e.g., grouping \textit{Films, Dramas, Songs, Books, etc.} in a ``\textit{Culture Product}'' cube and use one attribute such as \textit{genre} to distinguish them).

In each cube $\mathbb{K}_C$, the dimensions are derived from the structural primitives $(\mathcal{S}_C, \mathcal{A}_C, \mathcal{R}_C)$.
The set of subjects $\mathcal{S}_C = \{s_{C,1}, \dots, s_{C,J}\}$ forms the primary axis of the cube, and each instance (e.g., \textit{The Heart of Doreon, or Rolling in the Deep}) occupies a discrete coordinate along this axis.
Due to the diverse types of attributes and relations, we map each type of attribute $A \in \mathcal{A}_C$ and relation $R \in \mathcal{R}_C$ to its own orthogonal axis.
The total dimensionality of each cube is:
\begin{equation}
\text{Dim}(\mathbb{K}_C) = \underbrace{1}_{\text{Subject Axis}} + \underbrace{\sum_{i=1}^{M} \mathbf{1}(A^{(m)})}_{\text{Attribute Types}} + \underbrace{\sum_{j=1}^{N} \mathbf{1}(R^{(n)})}_{\text{Relation Types}},
\end{equation}
where $\mathbf{1}(\cdot)$ represents a mapping if the type is high-salience to a discrete orthogonal dimension.
Formally, the ontology-guide cube design for a class of subjects is represented as: 
\begin{equation}
    \mathbb{K}_{C} = \underbrace{\mathcal{D}_1}_{\text{Subject Dim}} \otimes 
    \underbrace{\mathcal{D}_1 \otimes \dots \otimes \mathcal{D}_G}_{\text{Attribute Dims}} \otimes 
    \underbrace{\mathcal{D}_{1} \otimes \dots \otimes \mathcal{D}_L}_{\text{Relation Dims}},
    \label{eq:ontology-guided-cube}
\end{equation}
where $\mathbb{K}_{c}$ denotes a cube for a class of subjects $C$, $\mathcal{D}_i$ is the $i^{th}$ dimension, and $1+G+L$ is the number of all cube dimensions, $G \le M$ and $L \le N$.



\subsection{Cube Construction }
\label{sec:cube_construct}
Following the design of the cube structure, we map documents to the corresponding cell(s), which then serve as a highly compressed yet semantically precise ``descriptor'' of the underlying content.
To achieve this, we employ a pre-trained LLM to extract instance values from each document across multiple dimensions of all cubes.
We prompt the LLM to go over all documents \textbf{only once} for the extraction process across all dimensions simultaneously.
This process is equivalent to tagging a document with labels across those dimensional values, which can be described as:
\begin{equation}
    \mathcal{E}(d) = \text{LLM}(d, \text{prompt}(\mathcal{D}_1,\dots,\mathcal{D}_{1+G+L})),
\end{equation}
where $\mathcal{E}$ represents the entity set extracted from a document $d$. The prompt with a structural schema is provided in Appendix \ref{app:prompt_ontology_schema}.

Then, the extracted entities are placed into the appropriate cube cells, and the associated document IDs are stored in their corresponding cabinets.
For example, Fig. \ref{fig:product_cube} illustrates that doc \#1 is allocated to a cube cell based on the associated label \lb \texttt{The Heart of Doreon, 1921, produced by} \rb. 
Once the cube is populated, it can be used \textit{repeatedly} to retrieve contextual documents for all questions.
It also supports \textit{incremental updates}, as the LLM only needs to process newly added documents to integrate their information into the established cube structure. 
Notably, specific dimensional values (e.g., \texttt{Release Date}) may remain null if the document does not contain the relevant information.
This highlights that the cube serves as a ``conceptual'' framework rather than a rigid physical array; a document may only partially instantiate the available dimensions. 
Consequently, unassigned dimensions do not consume storage space, ensuring the structure remains computationally efficient despite its multidimensionality.


\subsection{Cube-based Document Retrieval}
\label{sec:cube_rag}
Once cubes are designed and constructed, they function as a structured retriever within the RAG pipeline. Our cube structure retrieves relevant documents by navigating the dimensional values of the cube cells.
The retrieval process begins by decomposing a query into a set of entities and phrases aligned with the cube dimensions. 
\begin{equation}
    \{\mathcal{D}_1:e_1, \mathcal{D}_2: e_2, \ldots, \mathcal{D}_{1+G+L}: e_{1+G+L}\} = \text{LLM}_{\text{decomp}}(q),
\end{equation}
where $\{\mathcal{D}_i:e_i\}$ represents a pair of dimension and dimensional values. Noted that some dimensions remain empty if the query does not cover the related information. For simplicity, we use $\mathcal{E}(q)=\{e_1, e_2, \ldots, e_{l(q)}\}$ to denote the set of entities extracted from a query.

These decomposed components are then leveraged to search the relevant documents by comparing the extracted entities between the document and the query.
Our cube retrieval supports two matching strategies: \textit{sparse exact lexical match} and \textit{dense embedding match}.
The exact matching strategy is prioritized to use due to its high precision, scoring documents based on the frequency with which query entities appear:
\begin{equation}
    \text{score}(d, q) = \sum_{e_j \in \mathcal{E}(q)} \mathds{1}_{[e_j \in \mathcal{E}(d)]},
\end{equation}
where $\mathds{1}$ is an indicator function that equals 1 if an entity of the query exactly matches a term in the document, and 0 otherwise.

The semantic retrieval is achieved by computing the similarity between query entities and fine-grained labels along with cube dimensions.
When the similarity score exceeds a predefined threshold $\tau$, semantic retrieval is triggered. 
Both the query entities and fine-grained labels are projected into an embedding space, $\mathcal{Z}$, with an encoder. It can be represented as:
\begin{equation}
\begin{aligned}
    \mathcal{Z}(q) &= \text{Encoder}(\mathcal{E}(q)), \\
    \mathcal{Z}(d) &= \text{Encoder}(\mathcal{E}(d)), \\ 
    \text{score}(d, q) &= \text{sim}(\mathcal{Z}(q), \mathcal{Z}(d)),
\end{aligned}
\end{equation}
where $\text{sim}(\cdot)$ denotes a cosine similarity, $\mathcal{E}(q) = \{e_1, e_2, \dots, e_{l(q)}\}$ and $\mathcal{E}(d) = \{e_1, e_2, \dots, e_{l(d)}\}$ represent the decomposed entities of a query and document dimensional labels of cube cell(s).

\begin{figure}[ht!]
\centering
\includegraphics[width=0.99\columnwidth]{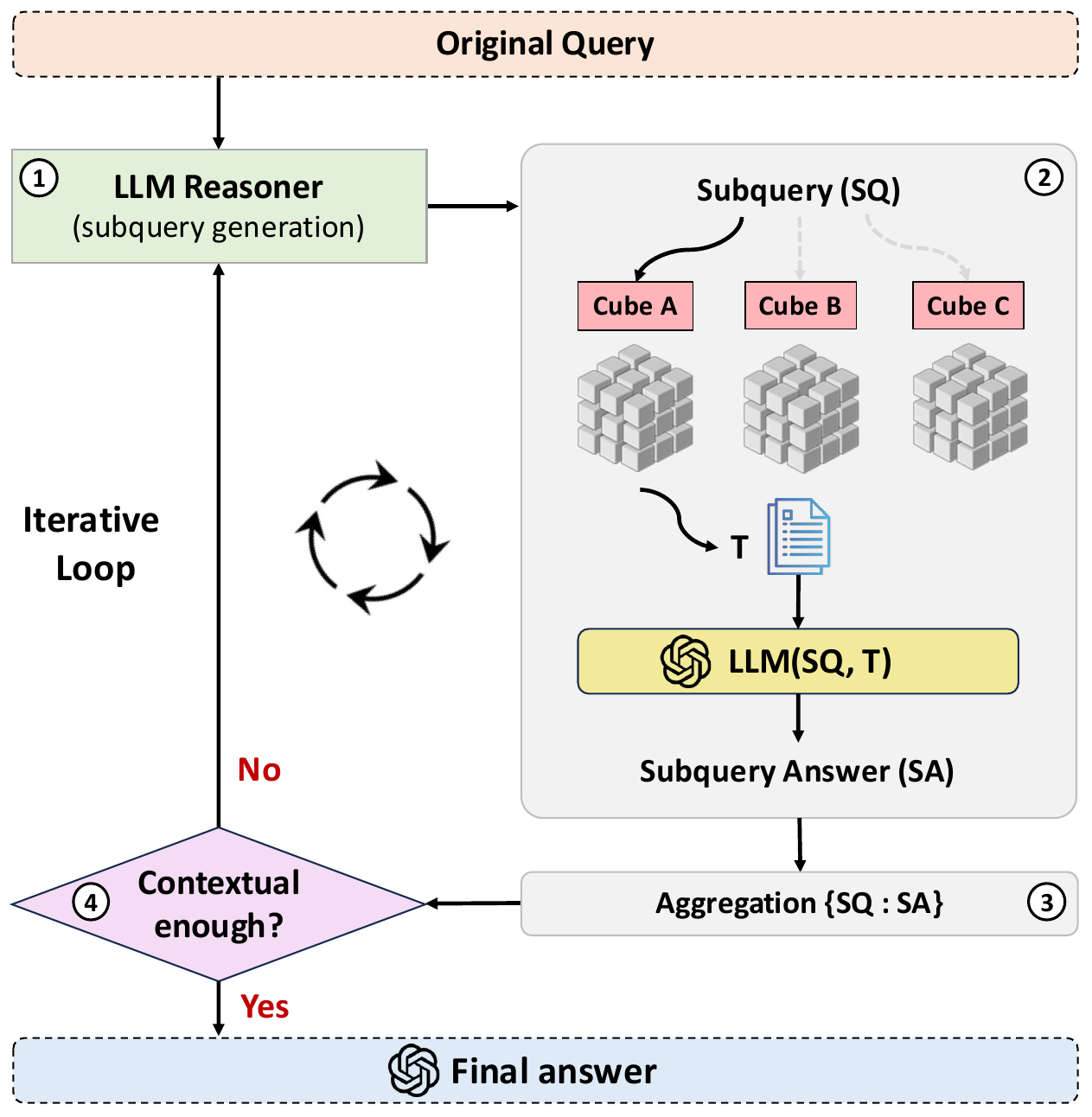}
\caption{\mcrag for Multi-turn Reasoning and Retrieval. 
    \tikzmarknode[mycircled]{a1}{1} Prompting LLMs to reason/generate subqueries; 
    \tikzmarknode[mycircled]{a1}{2} \mcrag for each subquery; 
    \tikzmarknode[mycircled]{a1}{3} Aggregating intermediate subqueries and their answers; 
    \tikzmarknode[mycircled]{a1}{4} Determining to output final answer or enter iterative reasoning and retrieval loop.
  } 
  \label{fig:framework}
\end{figure}

\subsection{\mcrag for Multi-step Reasoning and Retrieval}
\label{sec:mcrag}
We introduce \mcrag, a RAG framework centered on iterative multi-step reasoning and cube-based structured retrieval, tailored to solve multi-hop queries. 
The core idea is ``divide and conquer'': a complex multi-hop query is recursively decomposed into a sequence of one-hop subqueries. For each subquery, \mcrag dynamically selects the most appropriate specialized cube retriever to fetch contextually relevant documents. The resulting evidence is then used to generate an intermediate answer, which, in turn, informs the subsequent reasoning step and refines the search trajectory.
In the following sections, we detail the complete workflow (see Fig. \ref{fig:framework}), specifically focusing on the LLM-based reasoner for subquery generation, cube-based retriever selection, subquery answers aggregation, and iterative reasoning and retrieval loop.

\subsubsection{LLM Reasoner}
\label{sec:reasoner}
Given a multi-hop original query $q$, we prompt an LLM to generate one-hop subqueries (SQ) in multiple iterations.
In the initial iteration, the LLM is prompted only based on the original query to generate the logical subquery:
\begin{equation}
    SQ_1 = \text{LLM}_\text{subquery}(q, \text{prompt}_1).
\end{equation}
For the following iteration, $i (i > 1)$, the LLM is prompted based on the original query and the intermediate subqueries (SQ) and their corresponding sub-answers (SA) from \mcrag to continue reasoning by generating subsequent subqueries:
\begin{equation}
    SQ_i = \text{LLM}_\text{subquery}(q, \text{SQ}_{1:i-1}, \text{SA}_{1:i-1}, \text{prompt}_2).
\end{equation}

\subsubsection{Retriever Router}
Each subsquery may be related to different subjects (e.g., culture product, person, location), our \mcrag can flexibly select the most suitable subject-specialized cube for document retrieval. 
Given a subquery, $SQ$, an LLM with appropriate prompts is leveraged to decide which cube-based retriever should be activated. We provide all prompts in Appendix \ref{sec:prompt}.

\subsubsection{Intermediate Generation and Aggregation.}
Upon completion of the cube-based retrieval, the set of relevant documents $T$ is integrated with the corresponding subquery $SQ$ to construct an augmented prompt. This context-rich input enables the LLM generator to generate a subquery answer (SA). Formally, for each iteration $i$ within the reasoning loop, the process is defined as:
\begin{equation}
    SA_i = \text{LLM}_\text{sub\_QA}(\text{prompt}, SQ_i, T_i),
\end{equation}
where $i$ denotes the current iteration index. This intermediate answer $SA_i$ serves as the foundational evidence for the subsequent reasoning step, ensuring that the multi-hop trajectory remains grounded in the retrieved corpus.

Both intermediate subquery (SQ) and subquery answer (SA) pairs are aggregated together, $\{SQ_i, SA_i\}$, and serve as context for the subsequent reasoning and the final query answering.

\subsubsection{Multi-turn Reasoning and Retrieval Loop}
\label{sec:loop}
\mcrag is integrated into an iterative reasoning-and-retrieval loop designed to resolve complex multi-hop dependencies. The termination criterion, denoted as \texttt{``Contextual enough''} in Fig. \ref{fig:framework}, evaluates whether the accumulated set of subqueries and intermediate answers $\{SQ, SA\}_{1:i}$ provides a complete evidence chain to solve the original query. 
The reasoning loop will be executed until either the sufficiency condition is satisfied or the predefined iteration limit ($l$) is reached.


\subsubsection{Final Answer Generation}
Utilizing the accumulated evidence $\{SQ, SA\}_{1:i}$ alongside the original query, $q$, as context, the LLM generator is prompted to produce the final response:
\begin{equation}
    \text{Answer} = \text{LLM}_\text{final\_QA}(q, SQ_{1:i}, SA_{1:i}, \text{prompt}_3),
\label{eq:final_answer}
\end{equation}
where $i$ represents an intermediate iteration of the loop, and it is smaller than the iteration limit ($i \le l$).

\section{Experimental Setup}
\label{sec:exp}
\subsection{Datasets}
\label{sec:data}
Following the work \cite{gutierrez2024hipporag, gutierrez2025rag}, we evaluate \mcrag on four multi-hop QA benchmark datasets: \textbf{2WikiQA} \cite{ho2020constructing}, \textbf{HotpotQA} \cite{yang2018hotpotqa}, \textbf{MuSiQue} \cite{trivedi2022musique}, and \textbf{LV-Eval} \cite{yuan2024lv}, a challenging hotpotwikiqa mixup dataset including a small set of queries but a larger corpus. The data statistics are presented in Table \ref{tab:data_stats}.

\begin{table}[ht]
\centering
\caption{Dataset statistics. Length denotes the \# of words.}
\vspace{-2mm}
\resizebox{0.49\textwidth}{!}{
\begin{tabular}{l|cccc}
\toprule
\bf Datasets      & \# of docs          & \# of questions     & Question Length    & Answer Length    \\
\midrule
2WikiQA           & 6,119               & 1,000               & 4$\sim$27          & 1$\sim$14 \\
HotpotQA          & 9,811               & 1,000               & 5$\sim$42          & 1$\sim$14 \\
MuSiQue           & 11,656              & 1,000               & 6$\sim$40          & 1$\sim$18 \\
LV-Eval           & 22, 849             & 124                 & 6$\sim$31          & 1$\sim$8 \\
\bottomrule
\end{tabular}
}
\label{tab:data_stats}
\end{table}

\subsection{Baselines}
To assess the performance comprehensively, we select nine baselines across four categories:
1) \textbf{sparse retriever} BM25 \cite{robertson1994some}; 
2) \textbf{dense embedding} retrieval methods: Nvidia/NV-Embedv2 \cite{lee2024nv}; 
3) \textbf{structure-based} retrieval methods: RAPTOR \cite{sarthi2024raptor}, GraphRAG \cite{edge2024local}, HippoRAG \cite{gutierrez2024hipporag} and HippoRAG 2 \cite{gutierrez2025rag}; 
4) \textbf{prompt-based} iterative reasoning and retrieval methods: IRCoT \cite{trivedi2022interleaving}, RankCoT \cite{wu2025rankcot}.

\subsection{Evaluation Metrics}
We utilize multiple metrics to evaluate our method and baselines.
Following the work \cite{gutierrez2025rag, jin2025search}, exact match (EM) and F1 are used to assess the QA performance.

\subsection{Implementation Details}
We conduct experiments using two types of LLMs: GPT-4o-mini \cite{openai2024gpt4omini} and Llama3.3-70B-Instruct \cite{meta2024llama3} to generate responses.
The top-5 documents are retrieved as context for an LLM to generate the final answer. 
We prompt GPT-4o-mini as the LLM reasoner and cube router.
The prompts are provided in Appendix \ref{sec:prompt}.
We construct three cubes (cultural product, location, person) for 2WikiQA and LV-Eval datasets, four cubes (cultural product, location, person, politics) for MuSiQue dataset, and ten cubes (cultural product, location, person, politics, transportation, plant, machine, weather, organization, food) for HotpotQA dataset. 
The detailed cube dimensions are presented in Appendix \ref{sec:cube_dim}.
For the reproduced methods, we follow either their official GitHub repositories or the integrated GitHub repository\footnote{UltraRAG: https://github.com/OpenBMB/UltraRAG} to run the experiments using our datasets.
More reproducibility details for baseline methods are provided in Appendix \ref{sec:baseline}.
All experiments were conducted on a single H100 GPU with 80 GB memory.

\section{Experiments and Performance Results}
\label{sec:result}
\subsection{Accuracy}
Table \ref{tab:qa_perform} reports the EM and F1 results for QA performance. The observations and findings are as follows. 
First, we evaluate the different methods using two generators (GPT-4o-mini and Llama3.3-70B-Instruct) and the results show that \mcrag outperforms those state-of-the-art baselines on 2WikiQA, MuSiQue, and LV-Eval datasets, regardless of the LLM architectures. It shows the comparable performance on the HotpotQA dataset against the strongest baseline approach(es).
Second, we observe that NV-Embed-v2 and HippoRAG 2 show stronger performance than sparse lexical retriever (e.g., BM25) and iterative multi-step reasoning and retrieval approaches (e.g., IRCoT).
However, our \mcrag method, which belongs to the latter category, outperforms NV-Embed-v2 and HippoRAG 2, demonstrating the effectiveness of cube retrievers in resolving multi-hop queries.
Overall, it is $8.9\%$ over the average performance of various baselines, while the most improvement appears on the 2WikiQA dataset.

\begin{table}[ht!]
\centering
\caption{QA performance ($\%$) comparisons between our \mcrag and baselines. The best scores are in \textbf{bold} while the second-best scores are \ul{underlined}. }
\resizebox{\columnwidth}{!}{
\begin{tabular}{l|cc|cc|cc|cc}
\toprule
\multirow{2}{*}{\textbf{Methods}}  & \multicolumn{2}{c|}{\textbf{2WikiQA}}     & \multicolumn{2}{c|}{\textbf{HotpotQA}}   & \multicolumn{2}{c|}{\textbf{MuSiQue}} & \multicolumn{2}{c}{\textbf{LV-Eval}}   \\
\cmidrule(lr){2-3}  \cmidrule(lr){4-5}   \cmidrule(lr){6-7}  \cmidrule(lr){8-9}  
& EM      & F1                & EM      & F1                 & EM      & F1                  & EM      & F1         \\
\midrule

\rowcolor{gray!10}
\multicolumn{9}{l}{\textbf{\textit{Generator: GPT-4o-mini}}} \\
BM25 \cite{robertson1994some}                   & 46.9    & 51.8        & 49.1      & 59.8       & 17.9  & 27.3         & 6.5     & 9.4            \\
NV-Embed-v2 (7B) \cite{lee2024nv}               & 54.4 & 60.8           & \ul{57.3} & 71.0  & 32.8  & 46.0         & 7.3     & 10.0         \\
RAPTOR \cite{sarthi2024raptor}                  & 39.7 & 39.7           & 50.6      & 64.7       & 27.7  & 39.2         & 5.6     & 9.2          \\
GraphRAG \cite{edge2024local}                   & 45.7 & 61.0           & 51.4      & 67.6       & 27.0  & 42.0         & 4.9     & 11.0         \\
HippoRAG \cite{gutierrez2024hipporag}           & 59.4 & 67.3           & 46.3      & 60.0       & 24.0  & 35.9         & 4.8     & 7.6           \\
HippoRAG 2 \cite{gutierrez2025rag}              & \ul{60.5} & \ul{69.7} & 56.3      & \ul{71.1}  & \ul{35.0}  & \ul{49.3}      & \ul{10.5}    & \ul{14.0}  \\
IRCoT \cite{trivedi2022interleaving}            & 54.8 & 62.6           & 53.9      & 64.1       & 18.4  & 27.8         & 8.9     & 12.3        \\
RankCoT \cite{wu2025rankcot}                    & 52.5 & 62.7           & 53.5      & 65.8       & 23.8  & 35.8         & 5.6     & 8.2         \\
\mcrag \textbf{(Ours)}                   & \bf 63.2 & \bf 71.5   & \bf 57.5  & \bf 71.5       & \bf 39.5  & \bf 50.9 & \bf 11.3 & \bf 14.5      \\
\midrule

\rowcolor{gray!10}
\multicolumn{9}{l}{\textbf{\textit{Generator: Llama3.3-70B-Instruct}}} \\
BM25 \cite{robertson1994some}                   & 38.1 & 41.9           & 52.0      & 63.4       & 20.3     & 28.8     & 4.0     & 5.9         \\
NV-Embed-v2 (7B) \cite{lee2024nv}               & 57.5 & 61.5           & \bf62.8   & \ul{75.3}  & 34.7     & 45.7     & 7.3     & 9.8           \\
RAPTOR \cite{sarthi2024raptor}                  & 47.3 & 52.1           & 56.8      & 69.5       & 20.7     & 28.9     & 2.4     & 5.0          \\
GraphRAG \cite{edge2024local}                   & 51.4 & 58.6           & 55.2      & 68.6       & 27.3     & 38.5     & 4.8     & 11.2       \\
HippoRAG \cite{gutierrez2024hipporag}           & \ul{65.0} & \ul{71.8} & 52.6      & 63.5       & 26.2     & 35.1     & 6.5     & 8.4        \\
HippoRAG 2 \cite{gutierrez2025rag}              & \ul{65.0} & 71.0      & \ul{62.7} & \bf 75.5   & \ul{37.2}& \ul{48.6}& \ul{9.7}& \ul{12.9}     \\
\mcrag \textbf{(Ours)}                          & \bf 66.8 & \bf 74.3   & 61.3      & 72.9       & \bf 38.8 & \bf 49.7 & \bf 11.2& \bf 14.2     \\

\bottomrule

\end{tabular}
}
\label{tab:qa_perform}
\end{table}

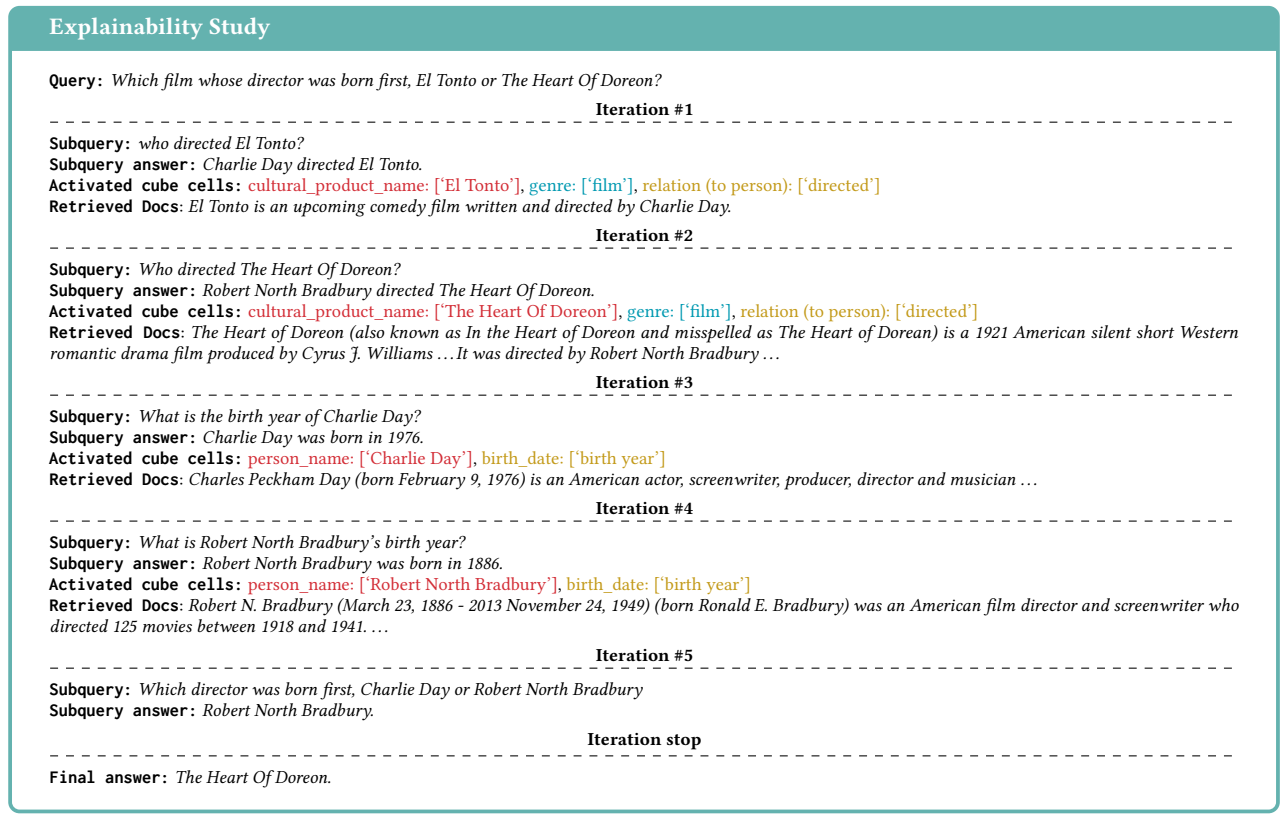
\begin{figure*}[ht!]
\makebox[0.95\textwidth][c]{%
\begin{tcolorbox}[colback=white,colframe=viridis3!70,title={\bf Explainability Study}, width=0.95\textwidth] 

\footnotesize
\textbf{\texttt{Query:}} \textit{Which film whose director was born first, El Tonto or The Heart Of Doreon?}
\begin{center}
    \vspace{-1em}
    \begin{tikzpicture}
      \draw[dashed] (0,0) -- (\linewidth,0)
      node[midway, above]{\textbf{Iteration \#1}};
    \end{tikzpicture}
\end{center}
\footnotesize
\textbf{\texttt{Subquery:}} \textit{who directed El Tonto?}  \\
\textbf{\texttt{Subquery answer:}} \textit{Charlie Day directed El Tonto.}  \\
\textbf{\texttt{Activated cube cells:}} \textcolor{red!90}{cultural\_product\_name: [`El Tonto']}, \textcolor{blue!90}{genre: [`film']}, \textcolor{yellow!90}{relation (to person): [`directed']}  \\
\textbf{\texttt{Retrieved Docs}}: 
\textit{El Tonto is an upcoming comedy film written and directed by Charlie Day.} 
\begin{center}
    \vspace{-1em}
    \begin{tikzpicture}
      \draw[dashed] (0,0) -- (\linewidth,0)
      node[midway, above]{\textbf{Iteration \#2}}; 
    \end{tikzpicture}
\end{center}
\footnotesize
\textbf{\texttt{Subquery:}} \textit{Who directed The Heart Of Doreon?}  \\
\textbf{\texttt{Subquery answer:}} \textit{Robert North Bradbury directed The Heart Of Doreon.}  \\
\textbf{\texttt{Activated cube cells:}} \textcolor{red!90}{cultural\_product\_name: [`The Heart Of Doreon']}, \textcolor{blue!90}{genre: [`film']}, \textcolor{yellow!90}{relation (to person): [`directed']}  \\
\textbf{\texttt{Retrieved Docs}}: 
\textit{The Heart of Doreon (also known as In the Heart of Doreon and misspelled as The Heart of Dorean) is a 1921 American silent short Western romantic drama film produced by Cyrus J. Williams \dots It was directed by Robert North Bradbury \dots} 
\begin{center}
    \vspace{-1em}
    \begin{tikzpicture}
      \draw[dashed] (0,0) -- (\linewidth,0)
      node[midway, above]{\textbf{Iteration \#3}}; 
    \end{tikzpicture}
\end{center}
\footnotesize
\textbf{\texttt{Subquery:}} \textit{What is the birth year of Charlie Day?}  \\
\textbf{\texttt{Subquery answer:}} \textit{Charlie Day was born in 1976.}  \\
\textbf{\texttt{Activated cube cells:}} \textcolor{red!90}{person\_name: [`Charlie Day']}, \textcolor{yellow!90}{birth\_date: [`birth year']}  \\
\textbf{\texttt{Retrieved Docs}}: 
\textit{Charles Peckham Day (born February 9, 1976) is an American actor, screenwriter, producer, director and musician \dots} 
\begin{center}
    \vspace{-1em}
    \begin{tikzpicture}
      \draw[dashed] (0,0) -- (\linewidth,0)
      node[midway, above]{\textbf{Iteration \#4}}; 
    \end{tikzpicture}
\end{center}
\footnotesize
\textbf{\texttt{Subquery:}} \textit{What is Robert North Bradbury's birth year?}  \\
\textbf{\texttt{Subquery answer:}} \textit{Robert North Bradbury was born in 1886.}  \\
\textbf{\texttt{Activated cube cells:}} \textcolor{red!90}{person\_name: [`Robert North Bradbury']}, \textcolor{yellow!90}{birth\_date: [`birth year']}  \\
\textbf{\texttt{Retrieved Docs}}: 
\textit{Robert N. Bradbury (March 23, 1886 - 2013 November 24, 1949) (born Ronald E. Bradbury) was an American film director and screenwriter who directed 125 movies between 1918 and 1941. \dots} 
\begin{center}
    \vspace{-1em}
    \begin{tikzpicture}
      \draw[dashed] (0,0) -- (\linewidth,0)
      node[midway, above]{\textbf{Iteration \#5}}; 
    \end{tikzpicture}
\end{center}
\footnotesize
\textbf{\texttt{Subquery:}} \textit{Which director was born first, Charlie Day or Robert North Bradbury}  \\
\textbf{\texttt{Subquery answer:}} \textit{Robert North Bradbury.}  
\begin{center}
    \vspace{-1em}
    \begin{tikzpicture}
      \draw[dashed] (0,0) -- (\linewidth,0)
      node[midway, above]{\textbf{Iteration stop}}; 
    \end{tikzpicture}
\end{center}
\textbf{\texttt{Final answer:}} \textit{The Heart Of Doreon.}

\end{tcolorbox}
}
\caption{Activated cubes and cube dimensions in \mcrag to represent the explainable retrieval process.}
\label{fig:case_study}
\end{figure*}

\subsection{Efficiency}
We investigate the efficiency of retrieval only and the whole RAG framework separately.

\subsubsection{Running time for retrieval only.} The retrieval efficiency is studied by computing the running time of different retrievers. 
For each dataset, 100 queries are selected for testing, and the average retrieval time is reported in Table \ref{tab:retrieval_efficiency}.
It shows that \mcrag achieves a significantly faster retrieval speed than \textbf{retrieval-heavy graph-based retrievers}, offering promising potential for large-scale, real-world retrieval tasks.
While \mcrag presents comparable retrieval efficiency compared with the sparse lexical retriever, BM25, and the dense embedding retriever, NV-Embed-v2, it delivers higher QA accuracy, as shown in Table \ref{tab:qa_perform}. 

\begin{table}[ht!]
\centering
\caption{Average running time ($s$) of retrieval only per query. $k$ refers to \# of documents. 
}
\vspace{-2mm}
\resizebox{0.999\columnwidth}{!}{
\begin{tabular}{l|cccc}
\toprule
              &\textbf{2WikiQA}  &\textbf{HotpotQA}  &\textbf{MuSiQue}  &\textbf{LV-Eval }  \\
\bf Methods   & ($k=6,119$)  &($k=9,811$)  &($k=11,656$)  &($k=22,849$)  \\
\midrule
BM25$_{(retrieval)}$              & 0.26   & 0.51   & 0.55    & 1.82      \\
NV-Embed-v2$_{(retrieval)}$       & 0.31   & 0.39   & 0.43    & 0.79      \\
GraphRAG$_{(retrieval)}$          & 0.44   & 1.67   & 4.46    & 11.82    \\  
HippoRAG 2$_{(retrieval)}$        & 3.75   & 5.33   & 3.58    & 4.86    \\      
\mcrag~$_{(retrieval)}$           & 0.15   & 0.33   & 0.76    & 0.41     \\
\bottomrule
\end{tabular}
}
\label{tab:retrieval_efficiency}
\end{table}

\begin{table}[ht!]
\centering
\caption{Average running time ($s$) of the framework per query. $k$ and $bs$ represent the \# of documents and batch size.
}
\vspace{-2mm}
\resizebox{0.999\columnwidth}{!}{
\begin{tabular}{l|cccc}
\toprule
              &\textbf{2WikiQA}  &\textbf{HotpotQA}  &\textbf{MuSiQue}  &\textbf{LV-Eval }  \\
\bf Methods   & ($k=6,119$)  &($k=9,811$)  &($k=11,656$)  &($k=22,849$)  \\
\midrule
BM25+Generator               & 0.79     & 0.97    & 1.12    & 2.30      \\
NV-Embed-v2+Generator        & 0.69     & 0.63    & 0.95    & 1.33      \\
GraphRAG                     & 0.44     & 1.67  & 4.46  & 11.82    \\  
HippoRAG 2                   & 4.24     & 5.69   & 5.61  & 9.37    \\  
\midrule
\mcrag (bs=5)                   & 1.37   & 3.37   & 1.73  & 2.12     \\
\mcrag (bs=10)                  & 0.71   & 1.95   & 0.91  & 1.31     \\
\bottomrule
\end{tabular}
}
\label{tab:all_efficiency}
\end{table}

\subsubsection{Running time for the whole RAG framework.}
Table \ref{tab:all_efficiency} summarizes the end-to-end inference latency for the RAG frameworks. For \mcrag, the total execution time includes subquery generation, cube selection, retrieval, and final answer generation. 
Compared to retrieval-heavy graph-based methods, we noticed that the primary determinant of \mcrag efficiency is the LLM calling on subquery generation and cube selection rather than retrieval. 
To reduce the latency in non-retrieval steps, we transition from sequential processing to asynchronous parallel batching. 
The last two rows reveal that framework latency significantly drops as the batch size ($bs$) increases, demonstrating its capacity for high-throughput scalability.


\subsection{Retrieval Explainability}
Given an original query, Fig. \ref{fig:case_study} shows a case study on the activated cube and cube dimensions for each iteration.
We notice that when the subquery asks film-related information, \mcrag will select the ``Culture Product'' cube and activate the cube cell(s) associated with the film attributes (\textit{film name, genre, and relation}). Moreover, the corresponding document(s) in the located cube cell(s) are retrieved as contextual information to solve the subquery.
Similar operations also happen to those queries that are related to other themes/subjects (e.g., person or location).
The retrieval explainability in our \mcrag ensures the retrieval process can be supervisable and verifiable, offering a significant and necessary feature for those high-stakes domain applications.

\subsection{Ablation Study}
We have different numbers of cubes for different datasets.
The results in Table \ref{tab:ablation} show that each cube plays a critical role in our \mcrag framework. 
For example, in the 2WikiQA dataset, the cultural product cube and person cube exert greater influence than the location cube, as most queries involve information about film directors, song composers, comparisons between films or persons, behaviors, relationships, and the nationality of persons, while only a small number of queries are related to location information.
The ablation studies on cube-based retrieval are presented in Appendix \ref{app:ablation} due to the limited space.

\begin{table}[ht!]
\centering
\caption{Ablation study by removing each cube. ``--'' represents that the cube was not built for that dataset. }
\vspace{-2mm}
\resizebox{0.999\columnwidth}{!}{
\begin{tabular}{l|cc|cc|cc|cc}
\toprule
\multirow{2}{*}{\textbf{Methods}}  & \multicolumn{2}{c|}{\textbf{2WikiQA}}     & \multicolumn{2}{c|}{\textbf{HotpotQA}}   & \multicolumn{2}{c|}{\textbf{MuSiQue}} & \multicolumn{2}{c}{\textbf{LV-Eval}}  \\
                & EM      & F1          & EM        & F1         & EM       & F1           & EM      & F1         \\
\midrule
w/o Cultural Product \cubeicon\  & 38.9    & 40.8        & 52.0      & 61.7       & 31.6     & 40.4         & 4.1     & 6.2        \\
w/o Location \cubeicon\          & 60.4    & 70.1        & 44.0      & 58.7       & 34.3     & 42.8         & 3.2     & 5.5        \\
w/o Person \cubeicon\            & 41.7    & 44.6        & 56.0      & 66.2       & 32.1     & 41.3         & 9.6     & 12.9       \\
w/o Politics \cubeicon\          & -       & -           & 54.0      & 66.7       & 37.4     & 49.9         & -       & -          \\
w/o Organization \cubeicon\      & -       & -           & 49.0      & 63.3       & -        & -            & -       & -          \\
w/o Plant \cubeicon\             & -       & -           & 50.0      & 57.1       & -        & -            & -       & -          \\
w/o Food \cubeicon\              & -       & -           & 55.0      & 64.6       & -        & -            & -       & -          \\
w/o Machine \cubeicon\           & -       & -           & 50.1      & 61.7       & -        & -            & -       & -          \\
w/o Weather \cubeicon\           & -       & -           & 51.1      & 61.4       & -        & -            & -       & -          \\
w/o Transportation \cubeicon\    & -       & -           & 52.0      & 61.8       & -        & -            & -       & -          \\
\midrule
\mcrag          & 63.2    & 71.5        & 57.3      & 69.4       & 39.5     & 50.9         & 11.3    & 14.5        \\
\bottomrule
\end{tabular}
}
\label{tab:ablation}
\end{table}
\section{Discussion}
\label{sec:discuss}

\subsection{Cube Storage and Construction Overhead}
Our cube is a multi-dimensional structure to represent the unstructured documents comprehensively and concisely. 
The comparison of entity index storage: \textbf{cube (120.3 MB) vs. HippoRAG 2 (831.1 MB)} for the 2WikiQA dataset, shows that the storage of cube structure is around 7 times smaller than dense graph-based RAG methods.
The underlying reasons are (1) the ontology-guided dimensions are highly salient for retrieval, rather than modeling numerous redundant and unnecessary pieces of information (entities and connections) in a dense graph; (2) the cube is a conceptual framework -- it takes no space if certain dimensional values are empty, even though there could be higher dimensions (similar to sparse ``array'' implementation). 
Therefore, the cube storage is memory-efficient. 

We leverage an LLM (GPT-4o-mini) to go over all documents for extracting ontology and dimensional values.
This process costs around \$24.01 for the cube constructions from 50,435 documents across four datasets.
Significantly, it can be \textit{repeatedly used} for answering various kinds of questions, once the cube construction is complete. 
Moreover, to deal with new documents, the cube is \textit{incrementally updated} efficiently since an LLM can go over those documents to extract dimensional information and update the established cube(s). 
Consequently, our cube construction is cost-efficient.

\subsection{Cube Design for Different Applications}
To apply the cube structure to other domains, we provide several suggestions as follows.
First, the hierarchical ontology structure needs to be extracted from documents.
Second, a single or multiple semantic classes $\{C_1, C_2, \ldots\}$ should be identified since they determine how many subject-specific cubes are required.
For each class, practitioners refine the relevant structural schema, $\Phi_C = (\mathcal{S}_C, \mathcal{A}_C, \mathcal{R}_C)$, ensuring that subjects ($\mathcal{S}$), high-salience attributes ($\mathcal{A}$), and relations ($\mathcal{R}$) are accurately captured. 
In the medical domain, for instance, this may involve identifying classes like ``Disease'' and ``Medication''. 
The schema $\Phi_{\text{Disease}}$ would then map condition names as subjects (e.g., Type 2 Diabetes), clinical markers as attributes (e.g., Symptoms like ``polyuria''), and therapeutic links as relations (e.g., ``Treated by'' [Medication]).
Third, these components are mapped into a multidimensional cube, facilitating granular, structured retrieval instead of standard text-based search. For complex multi-hop queries, such as identifying side effects for treatments of specific symptoms, the framework first locates the symptom coordinate in the ``Disease'' cube to identify the subject and its related medication. It then hops to the corresponding coordinate in the ``Medication'' cube to retrieve specific attribute values.


\subsection{Comparisons with Training-based Methods}
Table \ref{tab:qa_perform_vs_r1} compares our training-free \mcrag and recent reinforcement learning (RL)-based training methods.
We noticed that \mcrag outperforms \texttt{s3} on the 2WikiQA and MuSiQue datasets, while exhibiting a comparable performance on HotpotQA and LV-Eval datasets.
Moreover, the performance of \mcrag is close to that of (heavily trained) Search-R1 on 2WikiQA and LV-Eval datasets.
These results demonstrate the potential of our structural cube for information retrieval and question-answering without training. 
In the future, we aim to build more robust and powerful cube dimensions by leveraging minimal training effort.

\begin{table}[ht!]
\centering
\caption{QA performance ($\%$) comparisons between our training-free \mcrag and RL-based approaches.
LLM Generator: \texttt{Qwen2.5-7B-Instruct}.
}
\vspace{-2mm}
\resizebox{0.999\columnwidth}{!}{
\begin{tabular}{l|cc|cc|cc|cc}
\toprule
\multirow{2}{*}{\textbf{Methods}}  & \multicolumn{2}{c|}{\textbf{2WikiQA}}     & \multicolumn{2}{c|}{\textbf{HotpotQA}}   & \multicolumn{2}{c|}{\textbf{MuSiQue}} & \multicolumn{2}{c}{\textbf{LV-Eval}}  \\
\cmidrule(lr){2-3}  \cmidrule(lr){4-5}   \cmidrule(lr){6-7}  \cmidrule(lr){8-9}  
& EM      & F1                & EM      & F1                 & EM      & F1                  & EM      & F1           \\
\midrule

Search-R1 \cite{jin2025search}                  & 57.9 & 64.9           & 56.7  & 69.4      & 30.9 & 42.4       & 14.5   & 18.5   \\ 
s3 \cite{jiang2025s3}                           & 47.7 & 55.2           & 52.6  & 65.9      & 22.4 & 32.9       & 12.9   & 16.9   \\ 
\mcrag (Ours)                                   & 56.3 & 63.2           & 51.8  & 64.7      & 27.1 & 38.2       & 12.7   & 16.2   \\ 
\bottomrule

\end{tabular}
}
\label{tab:qa_perform_vs_r1}
\end{table}

\section{Conclusions}
\label{sec:conclude}
In this work, we analyze the limitations of existing RAG approaches on multi-hop queries. 
To address the limitations, we present an ontology-guided multi-dimensional structure (i.e., cube) for document representation and indexing.
With the cube design and construction, we propose \mcrag, a training-free RAG method tailored for the multi-hop questions, that consists of multiple subject-specific cubes for comprehensive and concise modeling of diverse subjects, attributes, and relations in documents. 
We further incorporate \mcrag into a multi-step reasoning and retrieval framework, which boosts the reasoning and retrieval capabilities and improves the generation performance.  
Experiments on four multi-hop QA datasets demonstrate that our \mcrag shows remarkable performance against state-of-the-art baselines.
Notably, \mcrag exhibits higher retrieval efficiency than graph-based RAG methods, proving its potential for scalable construction and applications. It also comes with inherent interpretable provenance in the retrieval process, enhancing the credibility and explainability in high-stakes domains.



\bibliographystyle{ACM-Reference-Format}
\bibliography{reference}

\appendix
\section*{Appendix}

\section{Prompt Templates}
\label{sec:prompt}

\subsection{Prompts for ontology extraction}
\label{app:prompt_ontology_schema}
\makebox[0.98\columnwidth][c]{%
\begin{tcolorbox}[colback=white,colframe=viridis3!80,title={\bf Prompt for ontology extraction}, width=0.98\columnwidth] %

\footnotesize

Given the following documents, you need to extract the ontology structure by extracting subjects and associated attributes and relations for each subject. \\

Please hierarchically output an abstract ontology-based schema in a two-level tree structure. For example, 

Level 1: Subject 1 \\
Level 2: Attribute 1, Attribute 2, Attribute 3, \ldots  \\
Level 2:  Relation 1, Relation 2, Relation 3, \ldots \\

You can group similar topics within a general subject, and you can group different types of attributes and relations. \\

\textcolor{red}{\{Documents:\}}

\end{tcolorbox}
}

\subsection{Prompts for query generation}
\label{app:prompt_query_generation}
\makebox[0.98\columnwidth][c]{%
\begin{tcolorbox}[colback=white,colframe=viridis3!80,title={\bf Prompt template for initial subquery}, width=0.98\columnwidth] %

\footnotesize
Given a multi-hop query, you MUST decompose it into several one-hop simple subqueries. \\ \\
Based on a query \textcolor{red}{\{original\_query\}}, what is the first, most direct and simplest single-hop question you need to ask to begin answering it? \\ 

You MUST begin with the simple ONE-HOP question in logic as simply as possible. Do NOT include an attributive clause in the question. \\ \\
You can not generate the subquery like ``What is the capital of the state where person A was born?'' since it is two hops.  \\ 

You must first ask where person A was born, then ask what the capital of the state. If the original query is asking who is younger, died earlier, etc. You need to start the first question with when? 



\end{tcolorbox}
}
\\  \\ \\
\makebox[0.98\columnwidth][c]{%
\begin{tcolorbox}[colback=white,colframe=viridis3!80,title={\bf Prompt template for intermediate subquery}, width=0.98\columnwidth] %

\footnotesize
Given the original query: \textcolor{red}{\{original\_query\}} and the intermediate steps taken so far: \textcolor{red}{\{intermediate\_answers\}}, What is the next logical question to ask to continue solving the original query? \\

You need to output the next logical question with one-hop as simple as possible. Do Not include attributive clause in the question. \\

If the original query can be fully answered with the current information, you must state 'FINAL ANSWER' directly without any other explanation. \\

For example, if the original query asks who is the someone's paternal grandfather, once intermediate steps include someone's father's father, then you should state 'FINAL ANSWER' directly without any other explanation. \\ \\
Otherwise, you must provide the next question only without any other explanations.

\end{tcolorbox}
}

\subsection{Prompts for cube router}
\label{app:prompt_router}
\makebox[0.98\columnwidth][c]{%
\begin{tcolorbox}[colback=white,colframe=viridis3!80,title={\bf Prompt template for 2wikimultihop dataset}, width=0.98\columnwidth] %

\footnotesize
Analyze the following query. If the query is centered on a cultural product (film, movie, song, magnize, band), e.g., who directed The Heart Of Doreon? You need to return ``CULTURAL\_PRODUCT'' since the key information is related to the film The Heart Of Doreon. \\

If the query is centered on a person, e.g., when was person A born or how old is the performer or director, you need to return ``PERSON'', since the key information is related to person A, the performer or director. \\

If the query is centered on location (asking specific places or geographic information), e.g., where was Marufabad and Nasamkhrali, you need to return ``LOCATION'' since the key information is related to the location Marufabad and Nasamkhrali. \\

Only output ``CULTURAL\_PRODUCT'' or ``PERSON'' or ``LOCATION''. \\

Query: \textcolor{red}{\{query\}}

\end{tcolorbox}
}

\subsection{Prompts for final output}
\label{app:prompt_final_output}
\makebox[0.98\columnwidth][c]{%
\begin{tcolorbox}[colback=white,colframe=viridis3!80,title={\bf Prompt template for the final answer}, width=0.98\columnwidth] %

\footnotesize
Analyze the following query. \\

Based on all the information gathered: \textcolor{red}{{intermediate\_answers}}, provide a final answer to the original query: \textcolor{red}{{original\_query}}. You must output the final answer without any other explanations. \\

If the query asks for dates or locations, only output the specific dates and locations. If the answer is a city, just output the city name and do not output the country it belongs to. \\

If it is a yes-or-no query, only output yes or no.  \\
If the query asks who, only output the person's name. \\
If the query asks for a comparison between two things, only output the one you think is correct without any other explanations. \\
If the query ask nationality of someone, directly output the country name.


\end{tcolorbox}
}

\section{Cube Dimensions}
\label{sec:cube_dim}

\subsection{Cultural product cube}
\begin{itemize}[leftmargin=0.6cm,nosep]
    \item \textbf{Culture Product name:} specific names, phrases on music, film, television, video game, literature, book, award, sport teams, sport (Football, Baseball, Basketball, Soccer, Olympics, Sports Leagues and Championships, Stadiums and Arenas, Sports Media and Broadcasting, Hockey, Rugby, Tennis, Golf, Motorsports, Cricket), entertainment, show, story, award \ldots
    \item \textbf{Product genre:} music, film, television, video game, literature, book, award, sport teams, sport, entertainment, show, story, award \ldots
    \item \textbf{Product relation (to Person):} released by, directed by, composed by, performed by, happened at, produced by \ldots
    \item \textbf{Product relation (to Location):} came out at, happened at, produced by \ldots 
    \item \textbf{Person name:} specific person names
    \item \textbf{Location name:} specific location names
\end{itemize}

\subsection{Location cube}
\begin{itemize}[leftmargin=0.6cm,nosep]
    \item \textbf{Location name:} specific names, phrases on school, university, river, body of water, bay, lake, mountain, hill, city, town, airport, bridge, park, reserve, historic site, historic landmark, company, church, \ldots
    \item \textbf{Location type:} school, university, river, body of water, bay, lake, mountain, hill, city, town, airport, bridge, park, reserve, historic site, historic landmark, company, church, \ldots
    \item \textbf{Location relation:} located at, is a part of, belongs to, next to, stands on, to the north of, lies at the foot of, historically in, \ldots
\end{itemize}

\subsection{Person cube}
\begin{itemize}[leftmargin=0.6cm,nosep]
    \item \textbf{Person name:} specific person names
    \item \textbf{Person role and occupation:} roles or occupations of a person (in the real world or the fictional character), such as director, composer, producer, writer, spouse, musician, actress, author, politician, scientist, founder, character, child, mother, father, grandfather, grandmother, father-in-law, stepmother, mother-in-law, stepmother, child-in-law, sibling-in-law, husband, wife, uncle, founder, paternal grandmother, maternal grandfather, \ldots
    \item \textbf{Person behavior:} die on, married to, work at, death, located at, wrote, awarded, won champion, produced, directed, live in, \ldots
    \item \textbf{Person nationality:} descriptions on nationality come from, is from, the country is, non-Native Americans, Alaskan natives, Swedish, \ldots
    \item \textbf{Relation (to Location):} affiliated with, born in, \ldots
    \item \textbf{Relation (to Product):} directed, composed, won award, \ldots
    
\end{itemize}

\subsection{Company and Organization cube}
\begin{itemize}[leftmargin=0.6cm,nosep]
    \item \textbf{Company and Organization name:} specific company and organization names
    \item \textbf{Relation (to Location):} location at, locate in, \ldots
    \item \textbf{Location name:} specific names
\end{itemize}

\subsection{Politics cube}
\begin{itemize}[leftmargin=0.6cm,nosep]
    \item \textbf{Politics name:} specific political events
    \item \textbf{Political event date:} specific dates related to political events
    \item \textbf{Political event location:} specific locations related to political events
\end{itemize}

\subsection{Plant cube}
\begin{itemize}[leftmargin=0.6cm,nosep]
    \item \textbf{Plant name:} specific plant names
    \item \textbf{Plant date:} specific dates related to plants, such as plant date, lifecycle, \ldots
    \item \textbf{Plant location:} specific locations related to plants, grow up at, live at, \ldots
\end{itemize}

\subsection{Transportation cube}
\begin{itemize}[leftmargin=0.6cm,nosep]
    \item \textbf{Transportation name:} specific transportation events
    \item \textbf{Political event date:} specific political dates
    \item \textbf{Political event location:} specific political events
\end{itemize}

\subsection{Machine cube}
\begin{itemize}[leftmargin=0.6cm,nosep]
    \item \textbf{Machine name:} specific machine events
    \item \textbf{Machine attributes:} service life, material, manufacturer, primary function \ldots
    \item \textbf{Machine relations:} designed by, manufactured by, shifted to, outfitted with, \ldots
\end{itemize}

\subsection{Weather cube}
\begin{itemize}[leftmargin=0.6cm,nosep]
    \item \textbf{Weather name:} specific weather events, such as wind, snow, precipitation, \ldots
    \item \textbf{Weather date:} specific weather dates related to weather events, \ldots
    \item \textbf{Weather location:} designed by, manufactured by, shifted to, outfitted with, \ldots
\end{itemize}

\subsection{Food cube}
\begin{itemize}[leftmargin=0.6cm,nosep]
    \item \textbf{Food name:} specific food names, such as beer, ice cream, candies, dessert, \ldots
    \item \textbf{Food attributes:} nutrition, ingredient, price, date, \ldots
\end{itemize}

\section{Reproducibility of Baselines}
\label{sec:baseline}
When running the three baselines (BM25, IRCoT, and RankCoT) with GPT-4o-mini as the generator, we adopted the UltraRAG~\citep{ultrarag2025} framework. This framework natively provides implementations of these baseline methods for text encoding, answer generation, and output evaluation. We further aligned the evaluation matrix to ensure that the scores produced under the UltraRAG setup were directly comparable to those from other implementations.

For Search-R1~\cite{jin2025search}, the retrieval corpus was reconstructed to be identical to the setting of \mcrag, ensuring a fair comparison. 
For retriever indexing, we employed \path{intfloat/e5-base-v2} to encode all corpus entries. In the LLM-based search and reasoning stage, we adopt \path{PeterJinGo/SearchR1-nq_hotpotqa_train-qwen2.5-7b-em-ppo-v0.3}. During each retrieval iteration, the retriever returns the top-2 ranked documents, with up to 5 iterative retrieval turns allowed per question.

\section{More Ablation Study}
\label{app:ablation}

As discussed in Sec. \ref{sec:cube_rag}, our cube-based retrieval supports the hybrid search strategies: \textit{sparse exact lexical match} and \textit{dense embedding match}. 
We conduct the ablation study on each search strategy below.
Table \ref{tab:ablation_em_sim} shows that both search strategies are significant.
 
\begin{table}[ht!]
\centering
\caption{Ablation study on sparse exact lexical match and dense embedding match. }
\vspace{-2mm}
\resizebox{0.999\columnwidth}{!}{
\begin{tabular}{l|cc|cc|cc|cc}
\toprule
\multirow{2}{*}{\textbf{Methods}}  & \multicolumn{2}{c|}{\textbf{2WikiQA}}     & \multicolumn{2}{c|}{\textbf{HotpotQA}}   & \multicolumn{2}{c|}{\textbf{MuSiQue}} & \multicolumn{2}{c}{\textbf{LV-Eval}}  \\
                & EM      & F1          & EM        & F1         & EM       & F1           & EM      & F1         \\
\midrule
w/o lexical match       & 47.9    & 51.8        & 55.4      & 65.9       & 31.6     & 46.4         & 6.5     & 9.2        \\
w/o embedding match     & 59.4    & 62.1        & 48.2      & 60.7       & 33.1     & 38.2         & 8.6     & 10.5        \\
\mcrag                  & 63.2    & 71.5        & 57.3      & 69.4       & 39.5     & 50.9         & 11.3    & 14.5        \\
\bottomrule
\end{tabular}
}
\label{tab:ablation_em_sim}
\end{table}

\end{document}